\documentclass[letterpaper, 10 pt, conference]{ieeeconf}  

\IEEEoverridecommandlockouts                              
\usepackage{graphicx}
\usepackage{amsmath}
\usepackage{amssymb}
\usepackage{siunitx}
\usepackage{times} 
\usepackage{balance}
\usepackage{subcaption}
\usepackage[font=small]{caption}     
\usepackage[maxfloats=256]{morefloats}
\makeatletter
\let\NAT@parse\undefined
\makeatother
\usepackage[bookmarks=false, linkcolor=blue, urlcolor=blue, citecolor=blue]{hyperref} 
\usepackage{fancyhdr}
\hypersetup{
     colorlinks=true,
     linkcolor=red,
     filecolor=magenta,      
     urlcolor=blue,
     pdfstartview={FitH},
     citecolor =blue
     }
\title{\LARGE \bf
 Hybrid Gripper Finger Enabling In-Grasp Friction Modulation Using\\Inflatable Silicone Pockets
}

\author{Hoang Hiep Ly, Cong-Nhat Nguyen, Doan-Quang Tran, Quoc-Khanh Dang, Ngoc Duy Tran*, Thi Thoa Mac, \\Anh Nguyen, Xuan-Thuan Nguyen, Tung D. Ta}

\begin{document}

\maketitle
\thispagestyle{empty}
\pagestyle{empty}

\begin{abstract}
Grasping objects with diverse mechanical properties, such as heavy, slippery, or fragile items, remains a significant challenge in robotics. Conventional grippers often rely on applying high normal forces, which can cause damage to objects. To address this limitation, we present a hybrid gripper finger that combines a rigid structural shell with a soft, inflatable silicone pocket. The gripper finger can actively modulate its surface friction by controlling the internal air pressure of the silicone pocket. Results from fundamental experiments indicate that increasing the internal pressure results in a proportional increase in the effective coefficient of friction. This enables the gripper to stably lift heavy and slippery objects without increasing the gripping force and to handle fragile or deformable objects, such as eggs, fruits, and paper cups, with minimal damage by increasing friction rather than applying excessive force. The experimental results demonstrate that the hybrid gripper finger with adaptable friction provides a robust and safer alternative to relying solely on high normal forces, thereby enhancing the gripper's flexibility in handling delicate, fragile, and diverse objects.
\end{abstract}



\section{Introduction}
In current robotic systems, ranging from industrial robots, biped robots to humanoid robots, the gripper is a key component that enables effective interaction between the robot and the surrounding environment~\cite{Shintake2018}. A gripper can serve not only as the end-effector to perceive target objects but also as the manipulator that handles them in specific tasks~\cite{Hughes2016}. The gripper performance directly affects the robot’s ability to accomplish complex operations. Therefore, the design of a flexible and high-performance gripper capable of manipulating a wide variety of objects with diverse shapes, sizes, and materials—while ensuring safety for users and preventing damage to fragile items—remains a challenge in the robotic field~\cite{Marco2013},~\cite{Achilli2020},~\cite{Billard2019}.

Most conventional industrial robots employ rigid grippers constructed from rigid links and joints (typically made of metal). These grippers are usually actuated directly by motors attached to the joints, or indirectly through cable/tendon-driven mechanisms~\cite{Liu2020},~\cite{Nishimura2018},~\cite{Do2023},~\cite{Ko2020},~\cite{Wei2023}. Thanks to their robust structure, rigid grippers provide high stability and repeatability, making them suitable for tasks requiring high precision and strength, such as grasping heavy objects in manufacturing~\cite{Long2020},~\cite{Samadikhoshkho2019}. However, their main drawback is limited adaptability to irregularly shaped objects. Moreover, when handling fragile or soft materials, rigid grippers may exert excessive gripping forces, potentially leading to damage the objects.
\begin{figure}[h!]
    \centering
    \includegraphics[width=\linewidth]{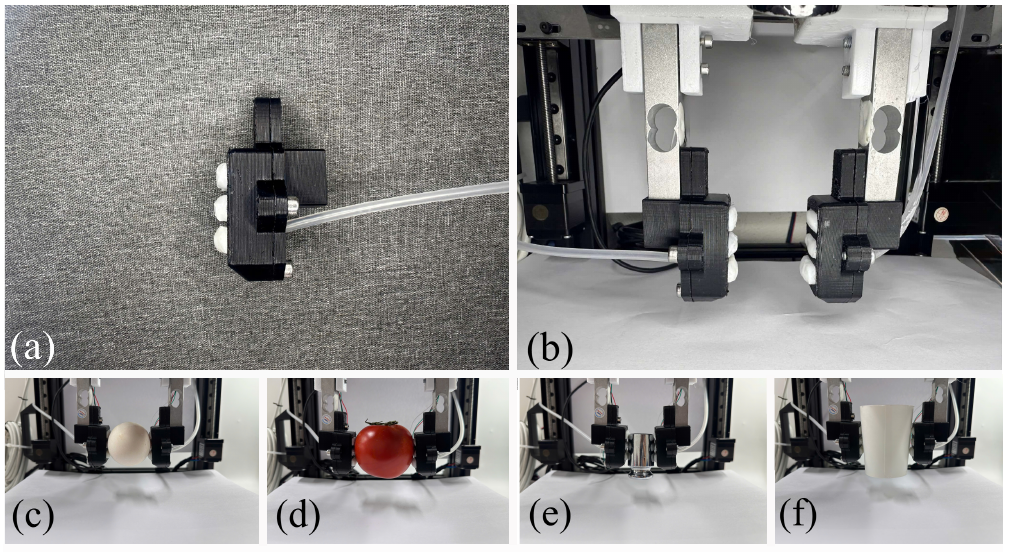}
    \caption{Hybrid robotic gripper for adaptive grasping using inflatable silicone pockets to grasp heavy and slippery objects, deformable objects, and various types of objects. (a) Gripper finger, (b) hybrid gripper system, (c-f) grasping various objects.}
    \label{fig:concept_overview}
\end{figure}

To overcome these limitations, soft grippers have been introduced using soft materials and compliant structures~\cite{Ham2018}. Soft grippers are highly deformable, allowing them to manipulate objects of varying shapes. The soft materials make these grippers safe for manipulating delicate/fragile objects and for human–robot interaction~\cite{Chen2021}. Grippers are mainly created from highly compliant materials such as silicone elastomers or hydrogels~\cite{Rus2015},~\cite{Majidi2019}. Various actuation technologies have been employed in soft grippers, including tendon/cable-driven mechanisms~\cite{Catalano2014}, shape memory alloy (SMA) wires~\cite{Li2024}, granular jamming~\cite{AboZaid2024}, and most commonly, pneumatic actuation~\cite{Deimel2016},~\cite{ilievski2011}. Despite these advantages (high flexibility, safety), soft grippers still have some limitations, such as weak grasping force, low response bandwidth, and limited controllability, compared to rigid grippers~\cite{Liu_2020}.

Driven by the demand for enhanced grasping performance, hybrid grippers—integrating both rigid and soft components—have recently attracted significant attention. By embedding soft elements into a rigid framework, hybrid grippers inherit the adaptability of soft grippers while retaining the strong gripping force and precise control of rigid grippers. For example, some designs use fluidic chambers for soft actuation combined with rigid linkages for structure~\cite{Park2018},~\cite{Park2020}. Another approach involves integrating soft materials only at the fingertips to increase contact area and compliance~\cite{Govindan2023}. A notable hybrid gripper employs pneumatic-based soft rings embedded at the finger joints to enable variable stiffness, allowing the fingers to bend backward and extend their grasping envelope~\cite{Tran2024}. This enables the stable picking up of thin and large objects, such as paper sheets or fabric panels. Pneumatic actuators are also used in pneumatic array systems to control the contact area between the device and the target objects~\cite{Pham2022-fm}.

Nevertheless, hybrid grippers, like rigid ones, still face a fundamental challenge: friction at the finger–object contact surface. When the surface friction coefficient is low, the gripper must apply a larger normal force to prevent slippage. The high force can lead to undesirable deformation or irreversible damage of fragile or deformable objects. A common solution is to attach high-friction materials (e.g., rubber or silicone) to the gripper finger surfaces to enhance grip performance. However, such fixed material provides only a constant friction level and lacks adaptability. Occasionally, excessively high friction can even reduce manipulation flexibility, making controlled sliding or precise release difficult. Although other approaches mix multiple materials geometrically to change the friction of the contact surface~\cite{Ta2022-kz, Ta2023-jp}, these are mainly applied to locomotion problems. Hence, there is a need for contact surfaces with tunable friction properties, enabling grippers to achieve high friction when firmly holding objects with low grasping force, and reduced friction when smooth release or controlled sliding is required.

To address these issues, this paper proposes a new hybrid gripper concept that integrates rigid fingers with a pneumatically actuated soft component designed to actively regulate contact conditions. By varying the internal pressure, the soft component can adjust both its stiffness and contact area, thereby adjusting the frictional interaction at the gripper finger–object in a controllable manner. The proposed design is expected to allow robots to grasp a wider of objects reliably, reduce the required gripping force, prevent slippage, and minimize damage to fragile objects. The remainder of this paper is organized as follows. First, the preliminary theoretical background on contact mechanics is introduced to underlying the proposed design. Next, the study presents the detailed mechanical structure and prototype of the hybrid gripper. In the following section, experimental evaluations, including measurements of the friction coefficient under different pneumatic pressures and grasping performance tests are revealed. Finally, the main contributions, limitations and directions for future research are summarized.

\section{Methods and Materials}

\subsection{Hybrid gripper finger in the context of contact theory}
\begin{figure}[tp]
    \centering
    \includegraphics[width=.9\linewidth]{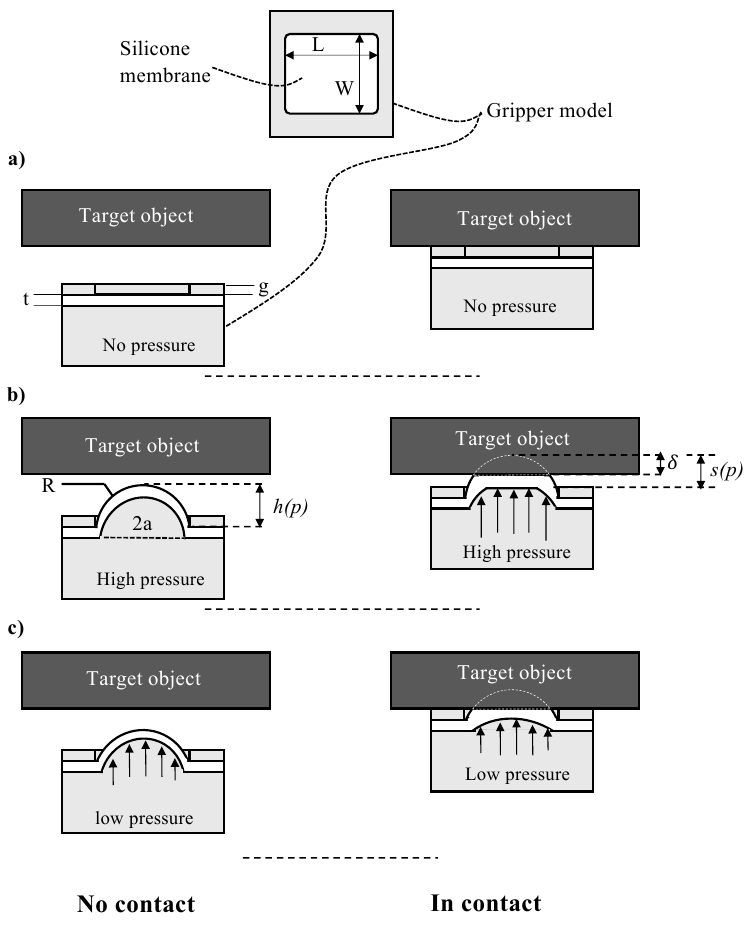}
    \caption{Modeling the contact mechanisms between a hybrid gripper finger and a target object.}
    \vspace{-1.5em}
    \label{fig:contact_mechanism}
\end{figure}

In this study, the contact phenomenon between a hybrid gripper finger and the surface of an object is modeled in a simplified manner, as shown in Fig.~\ref{fig:contact_mechanism}. Assume the gripper is a rigid box whose opening is a rectangle with dimensions of $W\times L$ and $(W<L)$, sealed by a soft silicone membrane of thickness $t$; the rim of the box is recessed by a gap $g$ below the plane of the metal target surface, as in Fig.~\ref{fig:contact_mechanism}. When an internal pressure $p$ is applied, the silicone membrane bulges into a spherical cap with apex deflection of $h(p)$. According to the bulge test~\cite{Xiang2005}, the relationship between $p$ and $h$ is presented as follows,
\begin{equation}
    p = 2\frac{\sigma_0 t}{a^2}h + \frac{4}{3}\frac{Et}{(1-\nu^2)a^4}h^3
\end{equation}
where $\sigma_0, a, E, \text{and } \nu$ denote the in-plane equibiaxial residual stress, the effective half-span ($0.5 W$), the Young’s modulus of the silicone membrane, and the Poisson’s ratio, respectively. For small $h$, $h \approx k_h p$ where $k_h = \frac{a^2}{2\sigma_0 t}$, therefore,
\begin{equation}
    h(p) = \min(k_h p, h_{\max})
\end{equation}
in which $k_h$ is the proportionality coefficient between bulge height and pressure, $h_{\max}$ is the limiting (critical) bulge height of the silicone membrane. Suppose the bulged portion forms a spherical cap; then the effective radius of curvature is presented as follows,
\begin{equation}
    R(p) = \frac{a^2+h(p)^2}{2h(p)}
\end{equation}
when $h(p) > g$ the protrusion beyond the rim can be computed as follow,
\begin{equation}
    s(p) = h(p) - g
\end{equation}
Because the pressurization tensions the membrane, an effective contact stiffness arises and is expressed as follows,
\begin{equation}
    E^*(p) = E_0(1+\eta p)
\end{equation}
where $E_0$ is the effective Young’s modulus at $p=0$ and $\eta$ is the proportionality factor describing how the effective stiffness scales with $p$. For a given normal load $N$, the contact area $A(N,p)$ and indentation $\delta(N,p)$ are given by the following equations,
\begin{equation}
    A(N,p) = \pi a_c^2
\end{equation}
\begin{equation}
    \delta(N,p) = \left[ \frac{3N}{4E^*(p)\sqrt{R(p)}} \right]^{2/3}
\end{equation}
where $a_c = \left( \frac{3NR(p)}{4E^*(p)} \right)^{1/3}$ is the contact radius of the bulged silicone against the object surface. Using Archard's Elastic Model of Friction~\cite{JFArchard1957} the friction force ($F_f$) is presented as follows,
\begin{equation}
    F_f(N,p) = \tau_s A(N,p) = \tau_s \pi \left( \frac{3R(p)}{4E^*(p)} \right)^{2/3} N^{2/3}
    \label{eq:friction_force}
\end{equation}
where $\tau_s$ is the interfacial shear stress for the silicone–object contact. Consequently, the friction coefficient of the silicone $\mu(N,p)$ against the object is
\begin{equation}
    \mu(N,p) = \tau_s \pi \left( \frac{3R(p)}{4E^*(p)} \right)^{2/3} N^{-1/3}
    \label{eq:friction_coef}
\end{equation}

With the above model, the silicone’s contact with the target plane can be categorized into three cases: no inner pressure (Fig.~\ref{fig:contact_mechanism}a), high inner pressure (Fig.~\ref{fig:contact_mechanism}b), and low inner pressure (Fig.~\ref{fig:contact_mechanism}c). In the absence of inner pressure—or when the pressure is too low so that $h(p) < g$ the friction is that between the rigid rim of the box and the target object, leading to a small friction force. When the pressure is increased so that $h(p) > g$ but remains insufficiently large, the appearance of a normal load $N$ pushes the bulged silicone downward; then $s(p) \le \delta$, and part of the box rim contacts the object, which reduces the gripper–object friction coefficient (Fig. ~\ref{fig:contact_mechanism}c). In contrast, when the inner pressure is high enough, for the same normal load $N$ the silicone fully contacts the object ($s(p) \ge \delta$), increasing the friction with the coefficient given by (\ref{eq:friction_coef}). In this case, the friction coefficient is clearly tunable via the inner pressure $p$. This is the theoretical basis for our proposed gripper, whose friction coefficient can be adjusted by controlling the inner pressure.

\subsection{Hybrid Gripper Finger}
\begin{figure}[bp]
    \centering
    \includegraphics[width=\columnwidth]{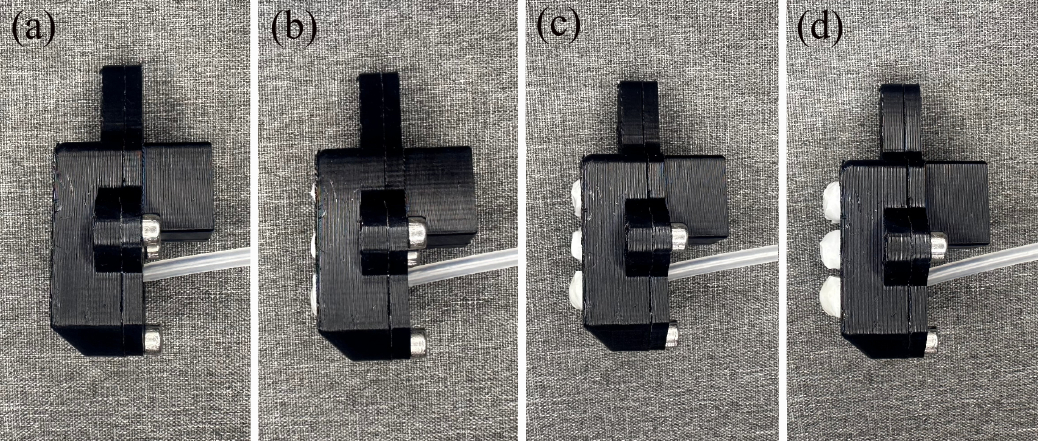}
    \caption{Inflation stages of the silicone air pocket at different pressure levels. (a) Total deflated, (b) slightly inflated, (c) inflated, (d) fully inflated.}
    \label{fig:inflation_stages}
\end{figure}

The finger comprises two components: a rigid outer shell that provides structural support during grasping, and a soft silicone air pocket housed inside the shell that inflates to increase friction when gripping an object. When air pressure increases in the air pockets, they expand outwards. At higher pressure levels, the pockets inflate to a greater degree, expanding until they reach the elastic limit of the silicone. These various stages of inflation are illustrated in Fig.~\ref{fig:inflation_stages}.





\subsubsection{Outer Shell}
\begin{figure}[tp]
    \centering
    \includegraphics[width=0.9\linewidth]{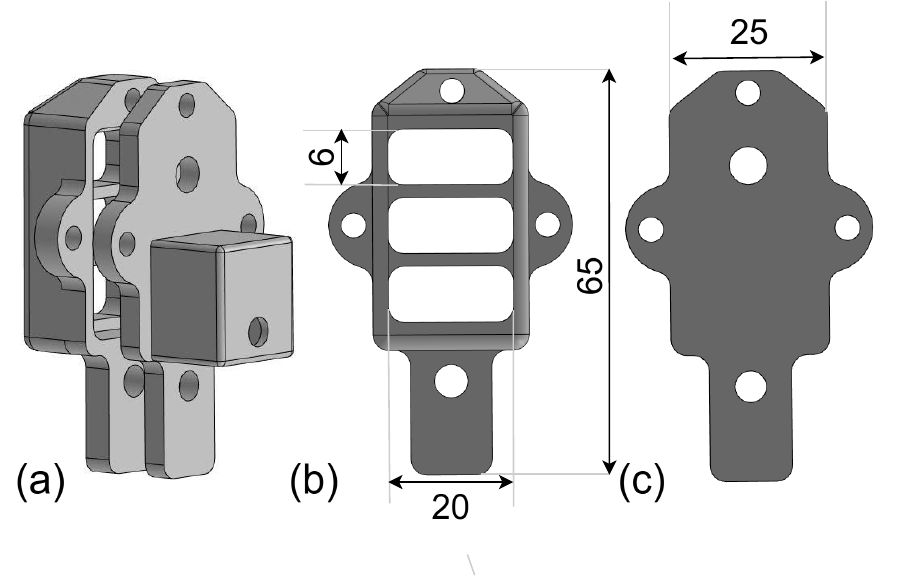}
    \caption{Design model of the outer shell. (a) Assembled view. (b) Front part with three grooves. (c) Back part with load-cell mount. Dimensions is in \SI{}{\milli\meter}.}
    \label{fig:outer_shell}
\end{figure}

Three grooves (cross-section of $20 mm\times 6 mm$) were added to the front surface of each finger, allowing the internal silicone pocket to expand and induce bending, as shown in Fig.~\ref{fig:outer_shell}. For ease of assembly, the finger was divided into two parts. The front part contains a hollow cavity to accommodate the silicone air pocket; the grooves enable outward expansion upon inflation. The back part  serves as a cover and includes a circular port for the air tube and mounting features for the load cells. The outer shell was fabricated by 3D printing using PETG.

\subsubsection{Silicone Air Pocket}
The air pocket is the soft component placed inside the outer shell. It was molded from silicone and designed to fit the internal geometry of the shell: a rectangular base with three rounded bulges on the front side, corresponding to the grooves of the outer shell. This design ensures a tight fit and allows the silicone to expand outward when pressurized. Compressed air is supplied through a silicone tube (inner diameter \SI{2}{\milli\meter}, outer diameter \SI{4}{\milli\meter}) inserted on the back side of the pocket.

\begin{figure}[htbp]
    \centering
    \includegraphics[width=\linewidth]{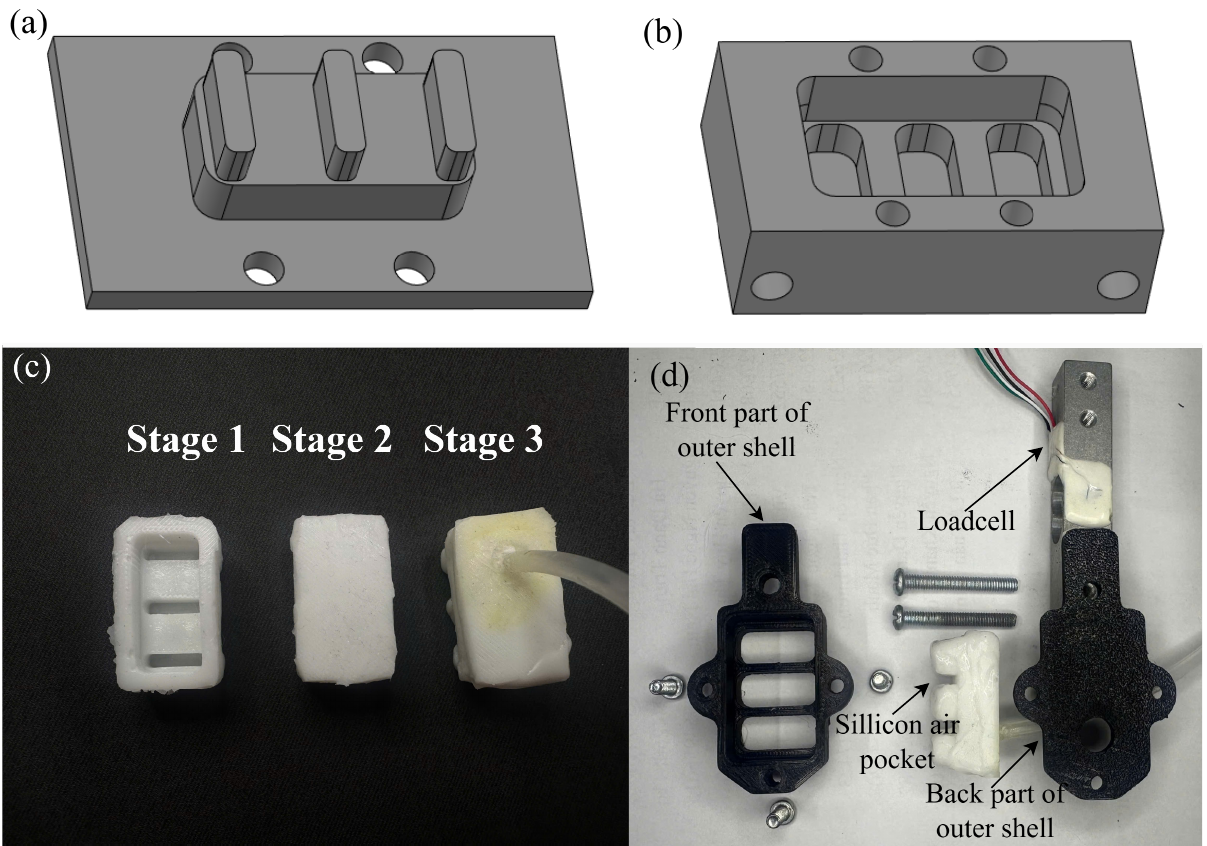}
    \caption{Fabrication of the silicone air pocket. (a) Female mold. (b) Male mold. (c) The three stages of fabrication. (d) The final product is assembled with a load cell mount.}
    \label{fig:silicone_mold}
\end{figure}

The air pocket was fabricated by molding. Both male and female molds were designed and 3D printed as illustrated in Fig.~\ref{fig:silicone_mold}(a) and~\ref{fig:silicone_mold}(b). The silicone (Shin Etsu KE-1416) was poured into the female mold, after which the male mold was pressed on top and fixed using screws through pre-designed holes. After curing, this process produced a silicone pocket with one open face. In the second step, another mold was used to seal the open surface, creating a closed air chamber. Finally, a small hole was punched in the back to insert the air tube. These three fabrication stages are illustrated in Fig.~\ref{fig:silicone_mold}(c).

\subsection{Gripper Prototype}
The silicone air pocket was inserted into the outer shell and sealed by fastening the two shell parts together with four screws. The air pressure generated from a pump was provide to the silicone pocket using air tubes, as shown in Fig.~\ref{fig:inflation_stages}. Two fingers were then mounted on one side of two $\SI{1}{\kilo\gram}$-load cell sensors, while the opposite side of each load cell was fixed to a support frame consisting of a linear guide and a sliding rail. A stepper motor (NEMA 17) was employed to actuate the gripper fingers. The motor converts rotational motion into linear motion of the two fingers, enabling both opening and closing actions. The entire gripper assembly was mounted onto a lifting system driven by two additional stepper motors with lead screws, allowing vertical motion of the gripper. In total, three stepper motors were used: one for finger actuation and two for the lifting system. These motors were controlled using an MKS~Gen~L~V1.0 control board.

\begin{figure}[htbp]
    \centering
    \includegraphics[width=\linewidth]{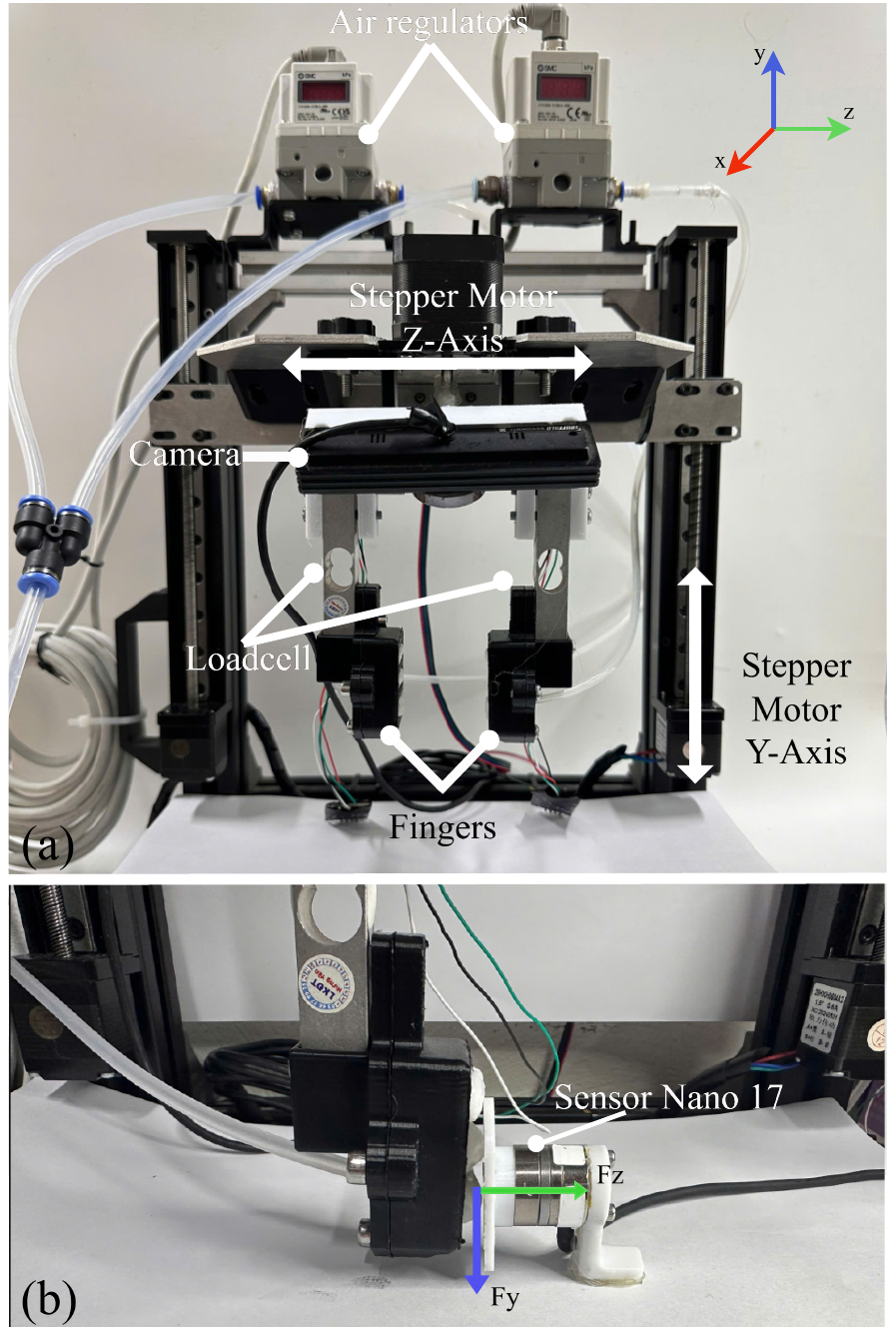}
    \caption{(a) Grasping experimental system using the hybrid gripper, (b)  Fundamental Experiment set up, the finger will move upward along Y-Axis\textcolor{red}{}}
    \label{fig:new_figure_3}
\end{figure}

For pneumatic actuation, compressed air was supplied by an air compressor and regulated using an ITV1030 electro-pneumatic valve. The valve input voltage ranged from \SIrange{0}{5}{\volt}, corresponding to an output pressure of \SIrange{0}{500}{\kilo\pascal}. The valve was driven by an Arduino Mega 2560, assisted by two Digital Analog converter-DAC modules (MCP4725) for voltage control. The complete experimental setup is illustrated in Fig.~\ref{fig:new_figure_3}(a).

\section{Experiments}

\subsection{Fundamental Experiment}
Fundamentally, an increase in applied pressure causes integrated air pockets to expand from the surface, resulting in a corresponding increase in frictional force, as mentioned in the previous section. To validate the hypothesis, this experiment utilizes a 6-axis force/torque sensor (ATI Nano17) and a data acquisition device (National Instruments). The sensor is mounted to a table with its Z-axis oriented perpendicularly to the finger to measure the normal force ($N$). At the same time, its Y-axis is aligned parallel to the finger's direction of motion, as shown in Fig.~\ref{fig:new_figure_3}(b). The procedure is conducted by pushing the sensor to apply a normal force ($F_{z}$), and then sliding the finger upwards across the surface. During this motion, the sensor records the tangential force ($F_{y}$) as the friction force, while $F_{z}$ represents the normal force, $N$. The coefficient of friction is calculated by the following equations,
\begin{equation}
\mu = \frac{F_{friction}}{N}=\frac{\left|F_{y}\right|}{\left|F_{z}\right|}
\end{equation}
Data is acquired from the moment the finger begins its upward slide until it is no longer in contact with the surface. All values of $F_{y}$  and $F_{z}$will be converted to their absolute values. The value of the coefficient of friction, $\mu$, is defined as the maximum value recorded during that time interval.

\subsection{Gripping Objects Experiments}
\subsubsection{Grasping Heavy and Slippery Objects Experiments}

The objective of this experiment is to evaluate the efficacy of the hybrid gripping mechanism in grasping heavy and slippery objects, a task that poses a significant challenge for conventional grippers. The underlying principle is that as the pressure within the integrated air pockets is increased, the silicone surface expands through apertures in the rigid shell. This action enlarges the effective contact area between the finger and the object, which in turn generates a higher frictional force and significantly enhances grasp stability.

The experiment utilized two steel masses (slippery payloads) of \SI{200}{\gram} and \SI{500}{\gram}. Emulating a human grasp, the gripper's fingers first approached the payload to ensure perpendicular contact with its parallel surfaces, then applied a compressive force before lifting. The experiment was conducted by investigating three constant normal force levels for each payload: \SIlist{3;3.5;4}{\newton} for the \SI{200}{\gram} object, and \SIlist{8;8.5;9}{\newton} for the \SI{500}{\gram} object. For each constant normal force, the internal pneumatic pressure was incrementally increased in steps: \SIlist{0;25;50;75;100;125}{\kilo\pascal}. At each (normal force, pressure) setpoint, 10 grasp trials were performed to determine the success rate. A trial was considered successful if the gripper lifted the steel mass clear of the support surface and maintained a stable hold for at least \SI{5}{\second}.

To lift the object, the total static friction force ($F_s$) generated by the gripper fingers must be greater than or equal to the object's weight ($W_{o}$).
\begin{equation}
    F_s \geq W_{o} = mg
\end{equation}
where $F_s$ is the total static friction force and $W_{o}$ is the gravitational force on the object.

To generate this friction, the gripper must apply a normal force ($F_z$) perpendicular to the object's surfaces. This force is measured directly by the integrated load cells. The static friction force is proportional to the normal force via the static coefficient of friction ($\mu_s$):
\begin{equation}
    F_{s} = \mu_s F_{z}
\end{equation}
where $F_z$ is the normal gripping force and $\mu_s$ is the static coefficient of friction between the fingertip and the payload.

From this, the minimum required normal force to prevent the object from slipping can be determined:
\begin{equation}
    F_z \geq \frac{mg}{\mu_s}
\end{equation}

\subsubsection{Grasping Deformable Objects Experiments}
In this experiment, paper cups were used as deformable test objects, with payloads of \SI{100}{\gram} and \SI{200}{\gram} placed inside to simulate loading conditions. The objective was to identify the grasping condition under which the cup undergoes minimal deformation. This was achieved by incrementally increasing either the gripping force or the inflation pressure, while holding the other parameter constant. For quantitative evaluation, a camera was mounted above the gripper to capture an image of the cup's rim after each grasping trial.

The degree of deformation was quantified by the Roundness Ratio\textbf{ ($R$)}, which compares the shortest (minor) diameter - $D_{\min}$  and longest (major) diameter - $D_{\max}$ of the object's rim post-grasp as follows,
\begin{equation}
    R = \frac{D_{\min}}{D_{\max}}
    \label{equa_roundness}
\end{equation}

$R$ value ranges from 0 to 1. $R$ value approaching 1 indicates that the minor and major diameters are nearly equal, signifying that the paper cup has maintained a nearly perfect circular shape with minimal deformation. Conversely, a smaller $R$ value (approaching 0) indicates a larger discrepancy between the diameters, signifying greater deformation.
\subsubsection{Grasping Different Objects Experiments}
\begin{figure}[htbp]
    \centering
    \includegraphics[width=\columnwidth]{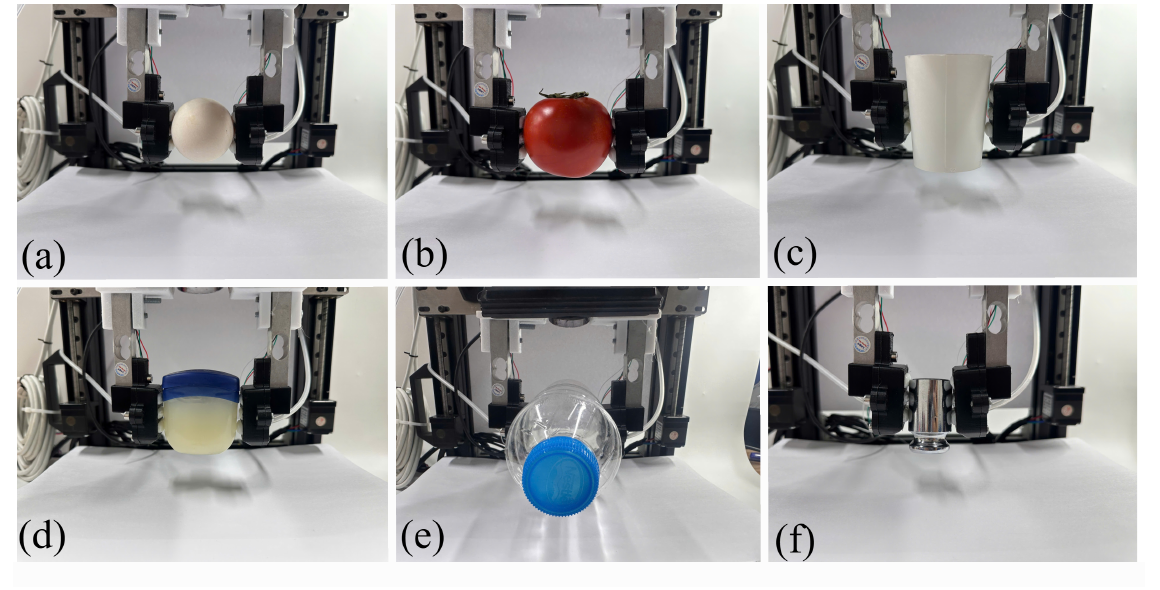}
    \caption{Grasping experiments with various objects. (a) Egg. (b) Tomato. (c) Paper cup. (d) Vaseline jar. (e) Water bottle. (f) Weight.}
    \label{fig:versatility_setup}
\end{figure}

In this experiment, a variety of objects with different shapes and surface properties were tested, including an egg, a tomato, a vaseline jar, and a plastic bottle, as shown in Fig.~\ref{fig:versatility_setup}. The purpose was to demonstrate the versatility of the gripper in handling a wide range of objects. Each object was grasped repeatedly under gradually increasing pressure levels, allowing us to evaluate the gripper’s adaptability and performance across diverse scenarios.

\section{Experiment Results}

\subsection{ Fundamental Experiment Results}
\begin{figure}[htbp]
    \centering
    \includegraphics[width=\columnwidth]{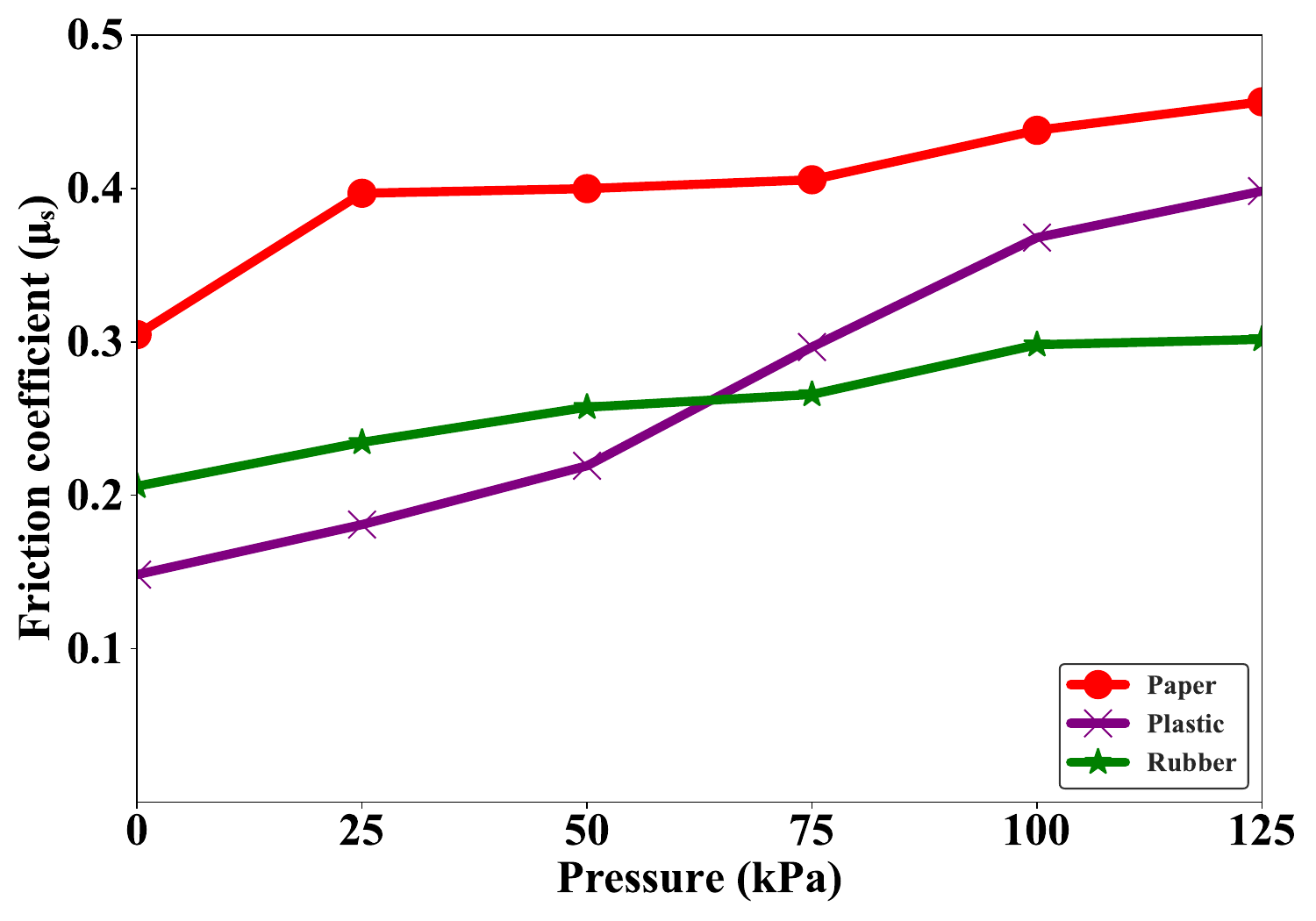}
    \caption{Relationship between coefficient of friction and applied pressure for different materials such as paper, plastic, and rubber.}
    \label{fig:friction_vs_pressure}
\end{figure}

Fig.~\ref{fig:friction_vs_pressure} illustrates the coefficient of friction ($\mu_{s}$) as a function of increasing applied pressure, with each material represented by a distinct curve. The results demonstrate a clear positive correlation: an increment in pressure consistently leads to an increase in the coefficient of friction.

This finding validates the hypothesis that at higher pressures, the integrated air pockets expand to increase the effective contact area. Consequently, the frictional force is augmented, enabling the gripping of heavier objects for a given grasping force. 

\subsection{Gripping Objects Experiment Results}

\subsubsection{Grasping Heavy and Slippery Objects Experiments Results}
\begin{figure}[htbp]
    \centering
    \includegraphics[width=\columnwidth]{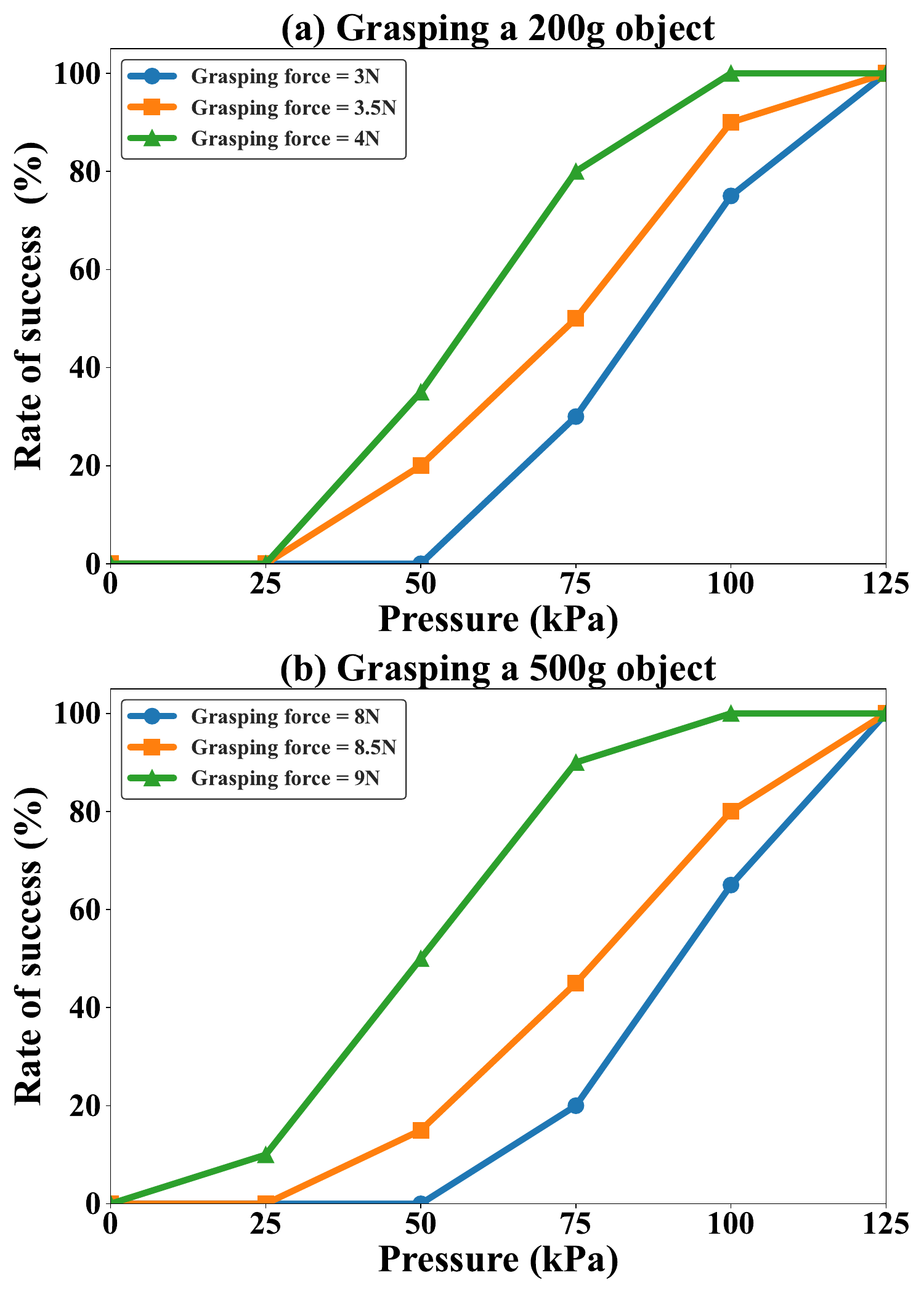}
    \caption{Grasping success rate as a function of pneumatic pressure for different normal force levels.}
    \label{fig:success_rate_plot}
\end{figure}

The grasping experiment for \SI{200}{\gram} and \SI{500}{\gram} steel mass was repeated 10 times. The grasping success rate for different grasping forces ($N$) and internal air pressure ($P$) are summarized in Fig.~\ref{fig:success_rate_plot}a and b, respectively. A strong positive correlation is evident across all trials: for a constant normal force, the success rate consistently increases as the pneumatic pressure rises. At \SI{0}{\kilo\pascal}, where only mechanical grasping is active, the success rate was minimal, underscoring the critical role of the pneumatic actuation in enhancing grasp stability.

These results provide evidence to validate the initial hypothesis. By holding $N$ constant, the observed transition from failed or slipping grasps to stable ones can be attributed solely to the variation of the actuation pressure $P$. This demonstrates that the effective coefficient of friction ($\mu_s$) is a controllable function of pressure. The physical mechanism might be that the expanded part of the silicone pocket increases the real contact area and creates micro-scale mechanical interlocking with the object's surface, thereby enhancing the friction coefficient. The results show that the gripper possesses a secondary modality for enhancing grasp stability—modulating its interface properties—rather than relying solely on increasing the normal force, a method that can be harmful to delicate objects.

\subsubsection{Grasping Deformable Objects Experiments Results}

\begin{figure}[htbp]
    \centering
    \includegraphics[width=\columnwidth]{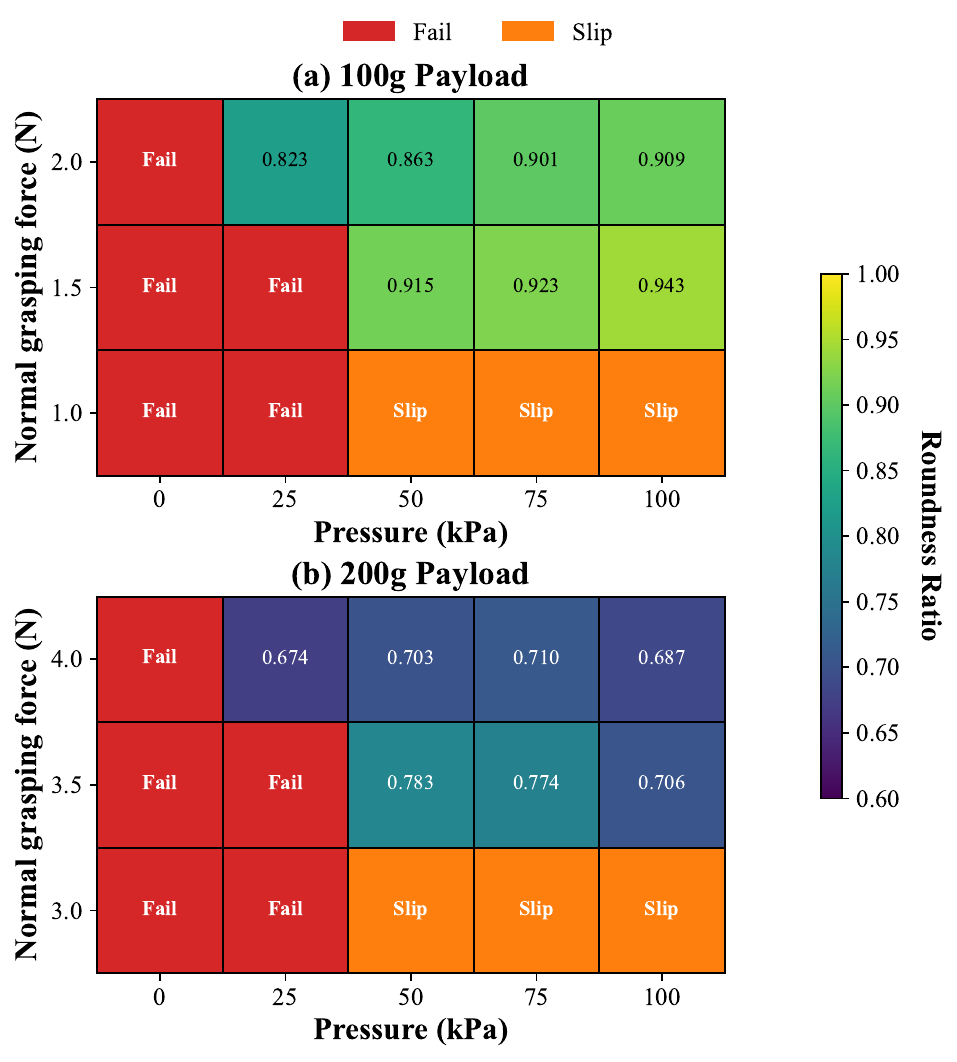}
    \caption{Analysis of deformable object grasping. The map identifies critical failure (red) and slippage (orange) regions, while the roundness ratio when successfully grasp of the cup rim is depicted by light green(maximum roundness)  to purple(minimum roundness)." }
    \label{fig:deformable_results}
\end{figure}

The quantitative results of the parameter sweep are visualized in the contour plots of Fig.~\ref{fig:deformable_results}. The experiment was conducted 10 times. These map the measured roundness ratio as calculated using equation~(\ref{equa_roundness}),  across the ($N, P$) parameter space. The color gradient visualizes the outcome of the roundness. The analysis reveals a distinct optimal status for each payload. For the \SI{100}{\gram} payload, the maximum roundness ratio $\approx 0.94$ is localized by $N \approx \SI{1.5}{\newton}$ and $P \approx \SI{100}{\kilo\pascal}$. As the payload increases to \SI{200}{\gram}, this optimal status shifts to $N \approx \SI{3.5}{\newton}$ and $P \approx \SI{50}{\kilo\pascal}$, achieving a best-case roundness ratio of $\approx 0.78$.

\begin{figure}[htbp]
    \centering
    \includegraphics[width=\columnwidth]{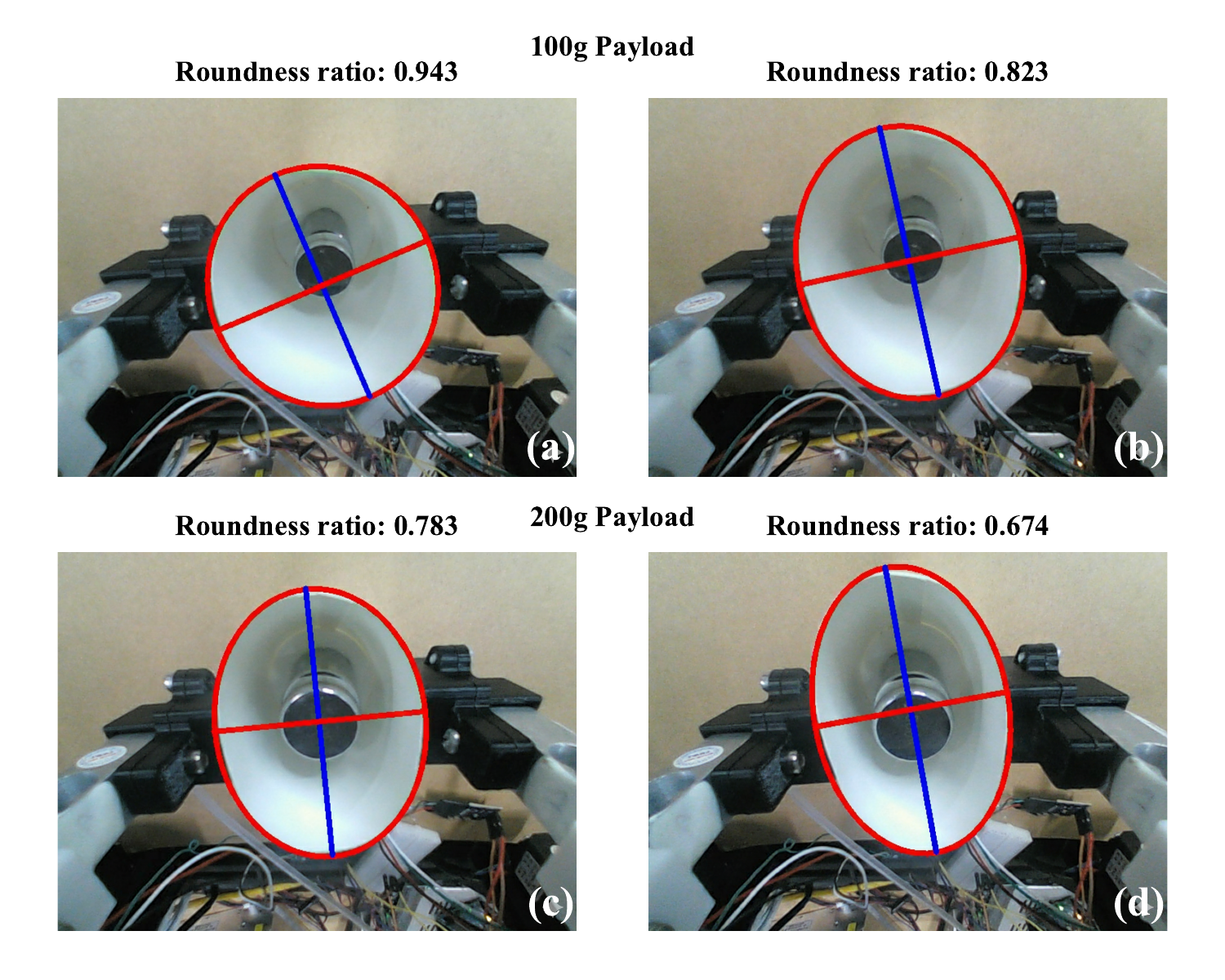}
    \caption{Visualization of the minimum and maximum deformation states of the paper cup. (a) Minimal deformation with a \SI{100}{\gram} payload. (b) Maximal deformation with a \SI{100}{\gram} payload. (c) Minimal deformation with a \SI{200}{\gram} payload. (d) Maximal deformation with a \SI{200}{\gram} payload.}
    \label{fig:new_figure_7}
\end{figure}

To reaveal the physical mechanism behind this performance, the gripper's morphological change was also quantified (Fig.~\ref{fig:new_figure_7}). The plot of the required initial opening versus actuation pressure provides direct evidence of the finger's pressure-induced volumetric expansion. Two key trends are evident: (1) for a constant pressure, a higher normal force requires a smaller initial opening; and (2) for a constant normal force, increasing the pressure requires a larger initial opening. This analysis confirms that the optimal, low-deformation grasps are achieved when the gripper maximizes its soft-body inflation to create a large, compliant contact area.

\subsubsection{Grasping Different Objects Experiment Results}
\begin{figure}[htbp]
    \centering
    \includegraphics[width=\columnwidth]{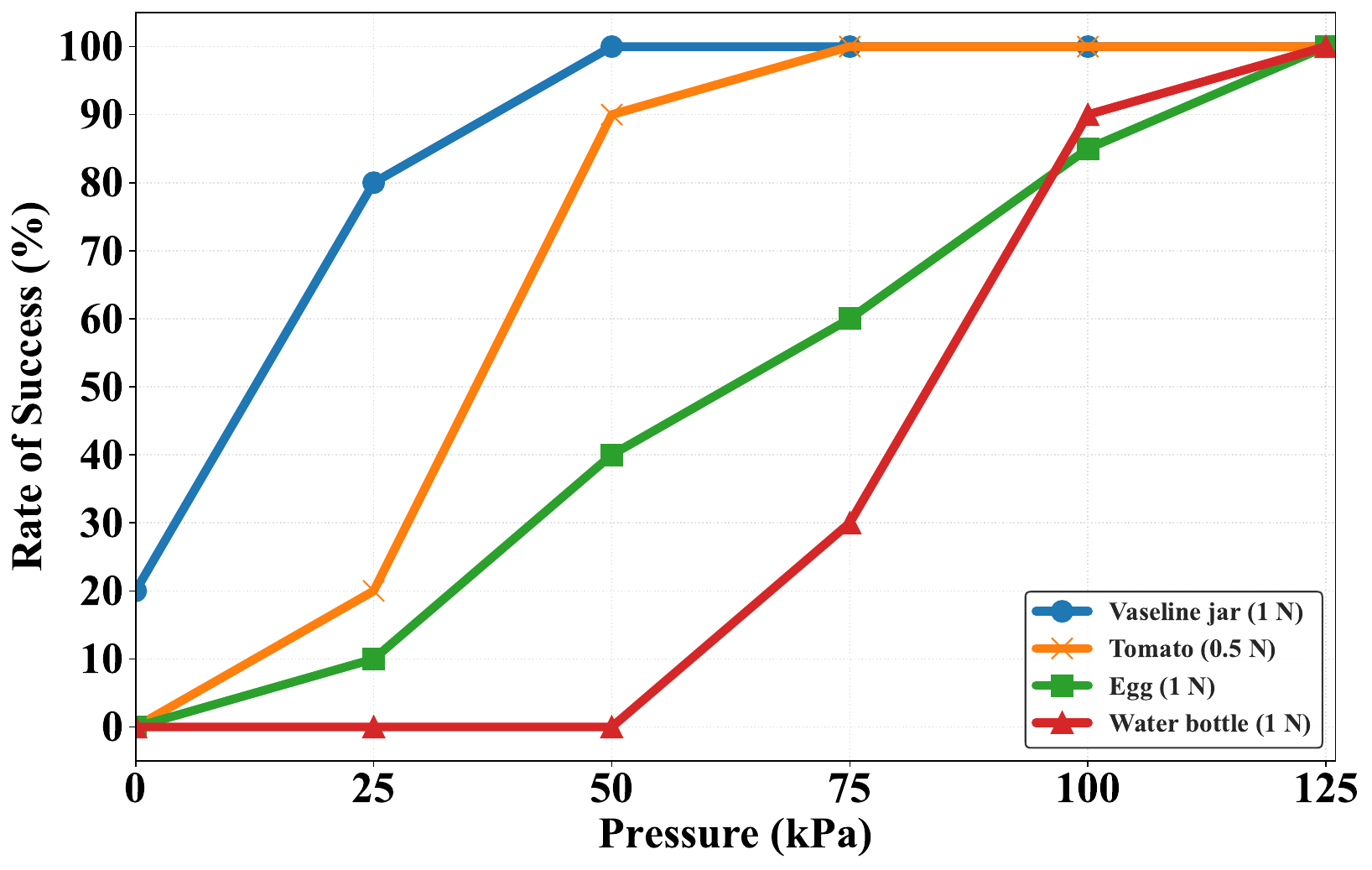}
    \caption{Grasping success rate as a function of actuation pressure for four different objects, each tested at a pre-determined normal force.}
    \label{fig:versatility_plot}
\end{figure}

Each object was grasped 10 times, and the grasping success rate for each of the four diverse objects is plotted as a function of actuation pressure in Fig.~\ref{fig:versatility_plot}. For all objects, the success rate demonstrates a strong positive correlation with the actuation pressure. At \SI{0}{\kilo\pascal}, the success rate is low, confirming that the normal force alone is insufficient for a reliable grasp, especially for slippery or heavy items. As the pressure increases, the silicone pads inflate, increasing the contact area and friction, which leads to a monotonic increase in the success rate.

The plots also reveal distinct performance curves for each object, highlighting the gripper's adaptability. For the rigid and slippery Vaseline jar, the success rate rises sharply, reaching 100\% at a low pressure of approximately \SI{25}{\kilo\pascal}. In contrast, the heavy and compliant water bottle required a higher normal force (\SI{3}{\newton}) and pressures exceeding \SI{80}{\kilo\pascal} to achieve near-100\% success. The fragile objects presented the greatest challenge; the egg required a pressure greater than \SI{50}{\kilo\pascal} to be grasped reliably, while the tomato's soft surface made achieving a consistently high success rate difficult, likely due to less effective mechanical interlocking. This experiment demonstrates the gripper's flexibility: by tuning both $N$ and $P$, it can adapt to diverse geometries, surface properties, and fragilities.

\section{Limitations \& Future Works}

The proposed hybrid gripper finger has demonstrated its effectiveness and safety in grasping various objects by adjusting the air pressure supplied to the soft component of the finger. However, the gripper finger still has several limitations. First, in this study, the soft pocket employs three bulges to regulate the gripper’s friction; however, the number and size of these bulges have not yet been optimized. In future work, we aim to optimize both the number and dimensions of the bulges to achieve higher gripping efficiency. Second, the current control of the gripper relies on the gripping force measured by a load cell. Nevertheless, additional forces also occur on the surface of the gripper finger, which makes precise control challenging. Moreover, in the grasping experiments, the internal air pressure of the gripper was manually preset for each task and was not automated. In future studies, we plan to integrate pressure sensors or flexible force sensors into the soft component to measure the contact force during operation, thereby improving control accuracy. In addition, feedback control or learning-based control algorithms will be applied to automate the grasping process, enabling the gripper to adapt more effectively to the target objects. Finally, the current gripper design is relatively simple and thus somewhat bulky; future work will focus on optimizing the design to make it more compact and compatible with widely used robotic arms.

\section{Conclusion}

In this study, we propose a hybrid robotic gripper finger that consists of a rigid structural shell combined with soft, inflatable silicone pocket. The gripper is capable of modulating surface friction during grasping by adjusting the internal air pressure of the silicone pocket. Fundamental experimental results indicate that increasing the internal air pressure enhances the friction between the gripper and the object. As a result, the gripper can achieve stable grasps on heavy objects with lower grasping force. This capability was demonstrated in experiments where the gripper successfully lifted heavy and slippery objects without increasing its gripping force. In addition, further experiments confirmed that the proposed gripper finger can grasp fragile and deformable objects effectively and safely. These findings highlight the potential of the hybrid gripper finger, with its pressure-controlled friction modulation, to optimize robotic interactions with diverse objects while ensuring their safety. In future work, we will focus on optimizing the gripper finger’s structure and implementing intelligent control algorithms to automate the grasping process.

\addtolength{\textheight}{-6cm}   
\bibliographystyle{IEEEtran}
\balance
\bibliography{biblio}

\end{document}